\documentclass[journal]{IEEEtranTIE}

\usepackage{graphicx}
\usepackage{stfloats}
\usepackage{amsmath,amssymb,mathtools,bm}
\usepackage{booktabs,multirow}
\usepackage[noadjust]{cite}      
\usepackage{siunitx}
\usepackage[caption=false,font=footnotesize]{subfig}
\usepackage{algorithm}
\usepackage{algpseudocode}
\usepackage{textcomp}
\usepackage{xcolor}
\usepackage[hidelinks]{hyperref}
\usepackage{makecell}
\usepackage{pifont}
\usepackage{etoolbox}

\makeatletter
\expandafter\patchcmd\csname\string\IEEEbiography\endcsname
  {\vskip \@IEEEBIOskipN plus 1fil minus 0\baselineskip}
  {\vskip \@IEEEBIOskipN}
  {}{\PackageError{main}{Unable to adjust biography spacing}{}}
\makeatother

\newtheorem{proposition}{Proposition}

\title{GPU-Accelerated Polygonal Signed Distance Functions for Real-Time Collision Avoidance}
\author{Taekwon~Ga and Jongeun~Choi%
\thanks{Taekwon Ga and Jongeun Choi are with the School of Mechanical Engineering,
Yonsei University, Seoul 03722, South Korea
(e-mail: taek111@yonsei.ac.kr; jongeunchoi@yonsei.ac.kr).}%
\thanks{Corresponding author: Jongeun Choi.}%
}

\hypersetup{pdfauthor={Taekwon Ga and Jongeun Choi}}

\newsavebox{\arxivnoticebox}
\sbox{\arxivnoticebox}{%
  \parbox[b]{\textwidth}{\centering\footnotesize
  This work has been submitted to the IEEE for possible publication.\\
  Copyright may be transferred without notice, after which this version may no longer be accessible.}%
}
\IEEEpubid{\usebox{\arxivnoticebox}}
\makeatletter
\edef\@IEEEpubidpullup{\the\dimexpr\ht\arxivnoticebox+\dp\arxivnoticebox+6pt\relax}
\makeatother

\begin{document}
\maketitle

\begin{abstract}
Optimization-based local planning and control require high-rate evaluation of collision-avoidance constraints over a prediction horizon. Accurately accounting for robot and obstacle geometry in these evaluations can be computationally expensive. The resulting bottleneck motivates collision-avoidance constraints that combine computational efficiency with geometric fidelity. The proposed polygonal signed distance function (PSDF) returns the minimum of exact signed distances between a convex polygonal robot footprint and convex obstacle components represented by their boundary edges. It is implemented as a training-free, branch-free tensorized geometric pipeline enabling batched GPU execution and automatic differentiation. The PSDF is embedded in model predictive control by locally linearizing the PSDF-based safety constraint within a sequential quadratic programming--based real-time iteration scheme, yielding the PSDF-embedded model predictive controller (PSDF-MPC). The design separates CPU/GPU computation so that the GPU evaluates batched PSDF values and gradients while the CPU solves a sparse quadratic program whose size and sparsity are determined by system dimensions and horizon length rather than obstacle and edge counts. Microbenchmarks show that PSDF scales favorably relative to geometric and learned collision-field baselines. Closed-loop comparisons in simulation, together with real-world navigation experiments, demonstrate that PSDF-MPC operates in real time and achieves collision-free navigation in dense polygonal environments.
\end{abstract}


\markboth{}{}

\section{Introduction}
\label{introduction}

Optimization-based local planning and control methods, particularly nonlinear model predictive control (NMPC), have become a standard paradigm for autonomous mobile robots because they handle system dynamics, reference tracking, and operational constraints within a single formulation \cite{seo2021nonaffine}. Collision avoidance is imposed through state constraints at every stage of the prediction horizon, requiring repeated evaluation within each control cycle. Their evaluation cost grows with the prediction horizon length, the number of obstacles, and the geometric complexity of the scene. In environments with high obstacle density or complex obstacle geometries, evaluating these constraints can therefore become a major computational bottleneck within each control cycle. Geometric fidelity is also essential in these environments, particularly in narrow passages where free space is limited. Conservative approximations of the robot footprint or obstacle boundaries can make traversable passages appear blocked, while models that omit parts of the actual geometry may fail to detect collisions.

\IEEEpubidadjcol

Existing collision representations face a trade-off between geometric fidelity and computational cost. Grid-based distance fields and simple geometric primitives provide fast queries but rely on discretized or simplified geometry \cite{oleynikova2017voxblox,hsieh2012nonlinear}. Duality-based formulations preserve polygonal geometry but introduce additional optimization variables and constraints whose number grows with obstacle count and geometric complexity \cite{zhang2018autonomous,zhang2020optimization,thirugnanam2022duality,thirugnanam2022safety,han2023rda,wu2024gpu}.

\begin{figure*}[t]
  \centering
  \includegraphics[width=\linewidth]{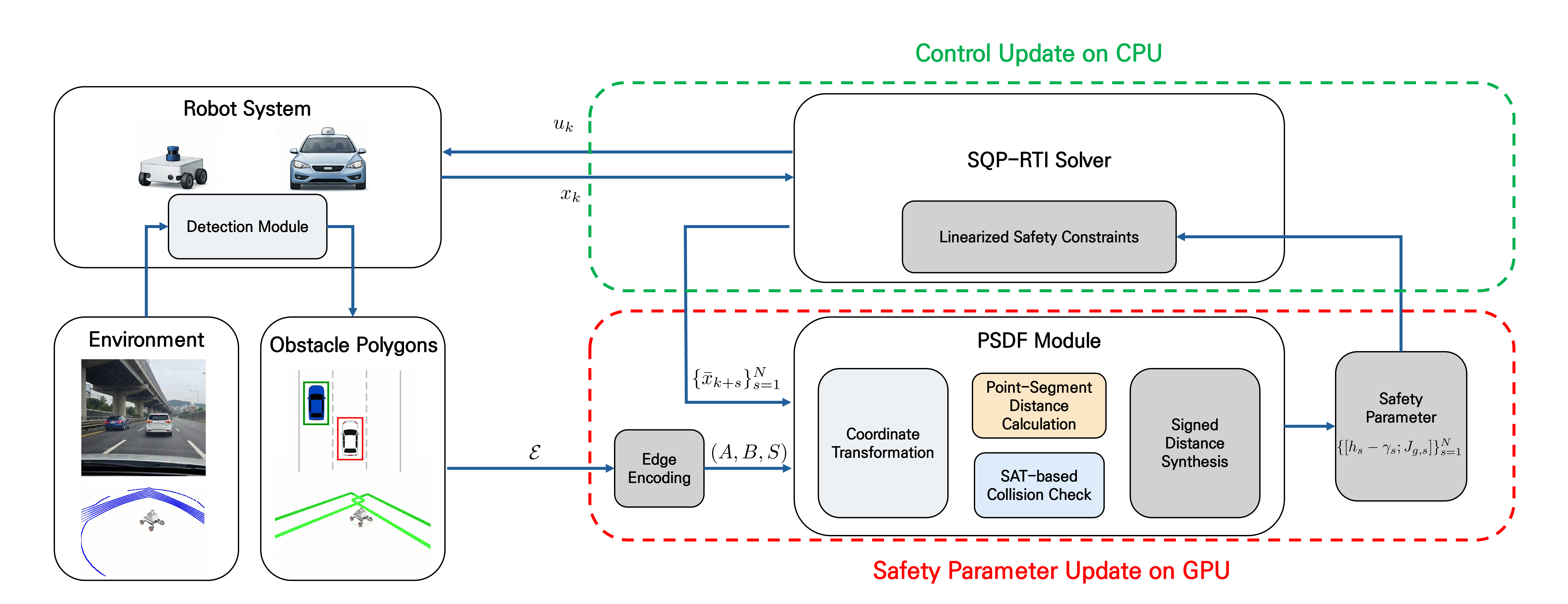}
  \caption{Architecture of the PSDF-embedded model predictive controller (PSDF-MPC) with separated CPU/GPU execution. Perceived obstacle polygons are encoded as padded edge tensors $(A,B,S)$. The GPU evaluates batched PSDF values and state gradients at the nominal predicted states to form coefficients for tightened linearized safety constraints. The CPU uses these coefficients in one SQP-RTI QP solve per control cycle to compute $u_k$.}
  \label{fig:arch}
\end{figure*}

These limitations motivate a geometry-exact signed distance function designed for efficient evaluation in real-time control. We introduce the polygonal signed distance function (PSDF), which combines point-to-segment distance calculations, separating axis theorem (SAT)--based penetration depth computation, and minimum reductions in a branch-free tensorized graph. This structure enables efficient batched GPU evaluation across obstacles and prediction stages. We integrate PSDF into model predictive control by locally linearizing collision-avoidance constraints within the sequential quadratic programming--based real-time iteration (SQP-RTI) scheme \cite{diehl2005real}, yielding PSDF-MPC.

The contributions of this paper are as follows.

\begin{enumerate}

\item \textbf{Exact signed-distance formulation.} An edge-based formulation that returns the minimum of exact signed distances between a convex polygonal robot footprint and convex obstacle components. The resulting PSDF is piecewise differentiable with respect to the robot state.

\item \textbf{Batched GPU evaluation and differentiation.} A branch-free tensorized implementation that evaluates PSDF values on the GPU and obtains state gradients through automatic differentiation, with geometric operations batched across obstacles and prediction stages.

\item \textbf{Geometry-independent quadratic program (QP) structure.} An SQP-RTI formulation that constructs one locally linearized safety constraint per constrained stage from PSDF values and gradients evaluated on the GPU. The CPU solves a QP whose size and sparsity remain independent of obstacle and edge counts.

\end{enumerate}

We evaluate the approach by benchmarking the computation of values and gradients used for collision avoidance against representative alternatives, comparing closed-loop navigation performance with optimization-based baselines, and conducting experiments covering onboard deployment and Ackermann parking. The PSDF module and the ROS packages used for the simulation and onboard experiments are available at~\url{https://github.com/Taek111/psdf_ros}.

\section{RELATED WORK}
\label{related_work}

\subsection{Simplified Collision Representations}

Optimization-based local planners commonly keep collision evaluation tractable by replacing the true geometry with a representation that enables efficient distance computation. Analytic primitives such as circles \cite{hsieh2012nonlinear} and ellipsoids \cite{rosolia2016autonomous} simplify collision constraints, but they can over-approximate robot footprints and become overly conservative in narrow passages. Grid-based distance fields, including costmaps and distance transforms \cite{maurer2003linear,felzenszwalb2012distance,oleynikova2017voxblox,han2019fiesta}, instead precompute distances on a fixed grid. Although queries are fast, narrow passages demand a fine grid at high memory and preprocessing cost, while coarse grids introduce discretization errors and nonsmooth gradients that can degrade solver conditioning.

\subsection{Geometry-Exact Collision Avoidance Constraints}

Retaining exact polygonal geometry inside the optimization requires either signed-distance evaluation or a smooth reformulation. Sequential convex trajectory optimization \cite{schulman2013finding} uses the Gilbert--Johnson--Keerthi (GJK) algorithm \cite{gilbert1988fast} and the expanding polytope algorithm (EPA) \cite{van2001proximity} for exact signed distances together with witness points and contact normals. These methods rely on iterative geometric queries, and evaluating all robot--obstacle pairs can become costly over a long prediction horizon. The resulting distance gradients can also be sensitive to numerical errors near degenerate contacts. Strong duality provides an alternative. Optimization-based collision avoidance (OBCA) \cite{zhang2018autonomous,zhang2020optimization} expresses the nonsmooth separation condition between convex bodies as smooth constraints in dual multipliers. The same principle underlies discrete-time control barrier functions (DCBF) for polytopes \cite{thirugnanam2022duality,thirugnanam2022safety} and accelerated variants based on the alternating direction method of multipliers (ADMM), such as RDA \cite{han2023rda}, or on GPU parallelism \cite{wu2024gpu}. Geometric fidelity is preserved, but every obstacle contributes its own multipliers and constraints, so the optimization problem inherits the complexity of the scene.

\subsection{Neural Models within Optimization}

Learning-based approaches use neural models for collision avoidance in optimization-based local planning and control. Implicit neural signed-distance representations are imposed directly as trajectory-optimization constraints \cite{Michaux-RSS-23}, while SDF-NMPC \cite{jacquet2025neural} encodes a single range image into a neural SDF under a spherical-robot assumption. NeuPAN \cite{han2025neupan} regresses strong-duality multipliers directly from point clouds in place of repeated dual solves, whereas NPField \cite{alhaddad2024neural} produces differentiable repulsive potentials from footprint and map encodings. These approaches offer efficient evaluation and differentiation, but their performance can be sensitive to the training distribution and may degrade under distribution shifts.

\section{Problem Statement}
\label{problem_statement}

\subsection{System and Environment}

Consider a robot with discrete-time dynamics given by

\begin{equation*}
x_{k+1}=f(x_k,u_k),\; x_k\in\mathcal X\subset\mathbb R^{n_x},\; u_k\in\mathcal U\subset\mathbb R^{n_u}, \tag{1}
\label{eq:rev1}
\end{equation*}

where $x_k$ and $u_k$ denote the state and the control input at time step $k$. The state $x_k$ contains the position and orientation of the robot. The admissible state and input sets $\mathcal{X}$ and $\mathcal{U}$ encode workspace bounds and actuator limits, respectively. We assume that $\mathcal{U}$ is compact and that $f:\mathbb{R}^{n_x}\times\mathbb{R}^{n_u}\to\mathbb{R}^{n_x}$ is continuously differentiable.

\textbf{Assumption 1.} The robot footprint and each obstacle component are convex, bounded polygons with nonempty interior. Their boundary edges have  positive length. Any nonconvex obstacle is represented as a finite union of convex polygons obtained by convex decomposition.

Let $P^{0}(x)\subset\mathbb R^{2}$ denote the robot footprint at state $x$, and let $\{P^{i}_{k}\}_{i=1}^{M_k}$ denote the obstacle polygons at time step $k$, where $M_k\in\mathbb Z_{+}$ is the number of obstacle components. The obstacle region is

\begin{equation*}
\mathcal O_k=\bigcup_{i=1}^{M_k}P^{i}_{k}. \tag{2}
\label{eq:rev2}
\end{equation*}

We omit the time index for simplicity and write each component in vertex form as

\begin{equation*}
P^i = \operatorname{co}\{v^i_{1},\dots,v^i_{m_i}\}, \tag{3}
\label{eq:rev3}
\end{equation*}

where $\operatorname{co}$ denotes the convex hull, $m_i$ is the number of vertices of $P^{i}$, and the vertices $v^{i}_{1},\dots,v^{i}_{m_i}$ are ordered counterclockwise.

For $p,q\in\mathbb R^{2}$, let $[p,q]:=\{(1-t)p+tq\mid t\in[0,1]\}$ denote the segment between them, so that $\partial P^{i}=\bigcup_{j=1}^{m_i}[v^{i}_{j},v^{i}_{j+1}]$ with $v^{i}_{m_i+1}=v^{i}_{1}$. Collision checking is performed against the environment edge set, defined as the disjoint union of the obstacle boundaries,

\begin{equation*}
\mathcal E_k:=\bigsqcup_{i=1}^{M_k}\partial P^{i}_{k}. \tag{4}
\label{eq:rev4}
\end{equation*}

The environment edge set $\mathcal E_k$ may change over time with perception updates. In our implementation and experiments, however, we hold $\mathcal E_k$ fixed over each prediction horizon.

\subsection{Signed Distance and the PSDF}

The signed distance between the robot footprint $P^{0}(x)$ and an obstacle polygon $P$ is defined as

\begingroup
\setlength{\jot}{6pt}
\begin{gather*}
\operatorname{sd}(P^{0}(x), P) := \operatorname{dist}(P^{0}(x), P) - \operatorname{pen}(P^{0}(x), P), \tag{5a}
\label{eq:rev5a} \\[2pt]
\operatorname{dist}(P^{0}(x), P) := \inf_{t} \{ \|t\| : (P^{0}(x) + t) \cap P \neq \emptyset \}, \tag{5b}
\label{eq:rev5b} \\
\operatorname{pen}(P^{0}(x), P) := \inf_{t} \{ \|t\| : (P^{0}(x) + t) \cap P = \emptyset \}, \tag{5c}
\label{eq:rev5c}
\end{gather*}
\endgroup

where $\operatorname{dist}(\cdot)$ and $\operatorname{pen}(\cdot)$ denote the minimum translation distances required to connect or separate the two sets, respectively. The signed distance is positive when the sets are separated, zero at contact, and negative when overlapping.

To aggregate the signed distances over all convex obstacle components, we define the polygonal signed distance function (PSDF), denoted by $\phi(x,\mathcal E)$, as their minimum:

\begin{equation*}
\phi(x,\mathcal E) \;:=\; \min_{i=1,\dots,M}\operatorname{sd}(P^{0}(x), P^i). \tag{6}
\label{eq:rev6}
\end{equation*}

The corresponding computation is developed in Section IV. Since the minimum is taken componentwise, \eqref{eq:rev6} applies unchanged to nonconvex obstacles represented as unions of convex polygons under Assumption 1, and the single scalar condition $\phi(x,\mathcal E)\ge 0$ ensures that the interiors of the robot footprint and the obstacle union do not intersect.

\subsection{Finite-Horizon Optimal Control Problem}

At each sampling instant $k$, given the measured state $x_k$ and a reference trajectory $\{x^{\mathrm{ref}}_{k+s}\}_{s=0}^{N}$, we solve

\begin{equation*}
\begin{aligned} \min_{\{u_{k+s}\}_{s=0}^{N-1}} \quad & \sum_{s=0}^{N-1}\Bigl( \bigl\|x_{k+s}-x^{\mathrm{ref}}_{k+s}\bigr\|_Q^2 + \bigl\|u_{k+s}\bigr\|_R^2 \Bigr) \\ & + \bigl\|x_{k+N}-x^{\mathrm{ref}}_{k+N}\bigr\|_{Q_f}^2 \\ \text{s.t.}\quad & x_{k+s+1}=f(x_{k+s},u_{k+s}), \\ & x_{k+s}\in\mathcal X,\quad u_{k+s}\in\mathcal U, \\ & h(x_{k+s+1},\mathcal E_{k+s+1})\ge 0, \\ & \forall s\in\{0,\dots,N-1\}, \\ & x_{k+N}\in\mathcal X. \end{aligned} \tag{7}
\label{eq:rev7}
\end{equation*}

Here $N$ is the prediction horizon length and $x_k$ is the initial condition of the prediction. The matrices $Q\succeq0$, $Q_f\succeq0$, and $R\succ0$ weight the stage tracking error, the terminal tracking error, and the control effort, respectively.

We define the PSDF-based safety function as $h(x,\mathcal E):=\phi(x,\mathcal E)-d_{\min}$, where $d_{\min}>0$ is the required minimum clearance. The safety constraint is

\begin{equation*}
h(x_{k+s},\mathcal E_{k+s})\ge0, \quad s=1,\dots,N. \tag{8}
\label{eq:rev8}
\end{equation*}

This constraint requires a clearance of at least $d_{\min}$ from the obstacle set at each constrained stage. The positive margin $d_{\min}$ excludes touching configurations and provides a practical buffer against perception and modeling errors and inter-sample motion.

Problem \eqref{eq:rev7} is generally nonconvex due to the nonlinear system dynamics and the nonconvex geometry of the collision-free state space. We solve the problem using SQP-RTI, as detailed in Section V.

\section{Polygonal Signed Distance Function}
\label{polygonal_signed_distance_function}

This section develops the evaluation of the polygonal signed distance $\phi(x,\mathcal E)$ defined in \eqref{eq:rev6}. Its value and state gradient are computed using a single branch-free computation graph with fixed tensor shapes. Fig.~\ref{fig:arch} illustrates the computational steps.

\subsection{Notation and Tensorized Environment Representation}

For a single pose, we omit the time index. Let $p\in\mathbb{R}^{2}$ and $\theta\in\mathbb{R}$ denote the translation and heading components of the full state $x\in\mathbb{R}^{n_x}$ in \eqref{eq:rev1}. The corresponding rotation matrix $R(\theta)\in\mathrm{SO}(2)$ maps the robot-local frame $\mathcal{L}$ to the world frame $\mathcal{W}$. The robot footprint in $\mathcal{L}$ is a fixed convex polygon, $\tilde{P}^{0}=\operatorname{co}(\tilde{V}^{0})$, whose $m$ vertices $\tilde{V}^{0}=\{\tilde{v}^{0}_{\ell}\}_{\ell=1}^{m}$ are ordered counterclockwise. At state~$x$, the corresponding footprint in the world frame is given by

\begin{equation*}
P^{0}(x)=\left\{R(\theta)v+p \;\middle|\; v\in\tilde P^{0}\right\}. \tag{9}
\label{eq:rev9}
\end{equation*}

To support branch-free batched GPU execution, we encode the environment in fixed-shape padded tensors. Let $\bar M\ge M$ and $\bar m\ge\max_i m_i$ be fixed capacities for obstacle components and edges per component, respectively, and write $\mathcal E^i:=\partial P^i$, so that $\mathcal E=\bigsqcup_{i=1}^M\mathcal E^i$. The segment endpoints are stored in fixed-shape padded tensors

\begin{equation*}
A,B\in\mathbb R^{\bar M\times\bar m\times 2}, \qquad S\in\{0,1\}^{\bar M\times\bar m}, \tag{10}
\label{eq:rev10}
\end{equation*}

in which $S$ is a validity mask such that $S^{i}_{j}=1$ identifies $[A^{i}_{j},B^{i}_{j}]$ as the edge $[v^{i}_{j},v^{i}_{j+1}]$ of $\mathcal E^{i}$, whereas $S^{i}_{j}=0$ indicates a padded entry that is excluded from all subsequent reductions by masking. The fixed bounds $(\bar M,\bar m)$ alone determine the shapes in \eqref{eq:rev10}, allowing obstacles with different vertex counts to form a uniform batch. A perception update changes only the values of $A$, $B$, and $S$, not their shapes or the associated computation graph.

At state $x$, the segment endpoints in the robot-local frame $\mathcal L$ are

\begin{equation*}
\begin{gathered} \tilde A^{i}_{j}(x)=R(\theta)^{\top}\!\left(A^{i}_{j}-p\right), \\ \tilde B^{i}_{j}(x)=R(\theta)^{\top}\!\left(B^{i}_{j}-p\right). \end{gathered} \tag{11}
\label{eq:rev11}
\end{equation*}

By expressing the obstacle geometry in the robot-local frame, we keep the footprint $\tilde P^{0}$ fixed and precompute its vertex and edge data. All subsequent computations therefore use $\tilde P^{0}$ and the localized segments $[\tilde A^{i}_{j},\tilde B^{i}_{j}]$. Unless indicated otherwise, the state argument $x$ is omitted from localized obstacle quantities below.

\subsection{Branch-Free Evaluation of Separation Distance}

The computational primitive for evaluating separation distances in the PSDF graph is the distance from a point to a line segment. This primitive is evaluated from every robot vertex to every obstacle edge and from every obstacle vertex to every robot edge, across all poses in a batch. To avoid branch divergence, the batched computation uses arithmetic, rectified linear unit (ReLU) operations, and masked reductions.

For a point $P\in\mathbb R^2$ and segment endpoints $A,B\in\mathbb R^2$, the orthogonal projection of $P$ onto the line through $A$ and $B$ is parameterized by a scalar $u$, which is clamped to $[0,1]$ using two ReLU operations.

\begin{equation*}
\begin{gathered} v=B-A,\qquad u=\frac{(P-A)^\top v}{\lVert v\rVert^{2}},\\ t=\operatorname{ReLU}(u)-\operatorname{ReLU}(u-1). \end{gathered} \tag{12}
\label{eq:rev12}
\end{equation*}

Assumption 1 ensures that $v\ne0$ for every valid segment. The clamp is piecewise linear with breakpoints at $u=0$ and $u=1$. The closest point on the segment and the point-to-segment distance are then

\begin{equation*}
Q=A+tv,\qquad d\bigl(P,[A,B]\bigr)=\lVert P-Q\rVert . \tag{13}
\label{eq:rev13}
\end{equation*}

Two disjoint convex polygons always admit a closest-point pair with at least one point at a vertex. We therefore evaluate vertex-to-edge distances in both directions. The first pool pairs robot vertices with obstacle edges,

\begin{equation*}
d_v^i=\min_{\ell,\;j:\,S_j^i=1} d\bigl(\tilde v^0_\ell,[\tilde A_j^i,\tilde B_j^i]\bigr). \tag{14}
\label{eq:rev14}
\end{equation*}

The second pool reverses the roles of the two bodies, measuring distances from obstacle vertices to robot edges:

\begin{equation*}
d_e^i=\min_{\substack{w\in\tilde V^i\\ \ell=1,\dots,m}} d(w,E_\ell), \tag{15}
\label{eq:rev15}
\end{equation*}

where $\tilde V^i(x):=\{\tilde A_j^i(x)\mid S_j^i=1\}$ is the obstacle vertex set in $\mathcal L$ and $E_\ell=[\tilde v^0_\ell,\tilde v^0_{\ell+1}]$, with $\tilde v^0_{m+1}=\tilde v^0_1$. The separation distance to obstacle $i$ is the minimum of the two pooled distances,

\begin{equation*}
d_{\mathrm{sep}}(x,\mathcal E^i)=\min\bigl(d_v^i,d_e^i\bigr). \tag{16}
\label{eq:rev16}
\end{equation*}

It is strictly positive while $P^0(x)$ and $P^i$ are disjoint and decreases to zero at contact. Because the two pools contain a minimizing pair, $d_{\mathrm{sep}}(x,\mathcal E^i)$ equals the Euclidean distance $\operatorname{dist}(P^0(x),P^i)$ exactly under disjointness.

\subsection{SAT-Based Overlap Detection and Penetration Depth}

The separation distance $d_{\mathrm{sep}}(x,\mathcal E^i)$ is sufficient when the footprint and an obstacle are disjoint. To handle contact and overlap, we apply the separating axis theorem (SAT) to detect overlap and quantify penetration, using the unit edge normals of both polygons as candidate separating axes. For each valid obstacle edge, define

\begin{equation*}
u_j^i=\tilde B_j^i-\tilde A_j^i,\qquad n_j^i=\frac{\operatorname{rot90}(u_j^i)}{\lVert u_j^i\rVert}, \tag{17}
\label{eq:rev17}
\end{equation*}

where $\operatorname{rot90}(x,y):=(-y,x)$ and the denominator is nonzero by Assumption 1. Together with the unit normals $n_\ell^0$ of the robot edges $E_\ell$, these form the candidate axis set

\begin{equation*}
\mathcal N^i=\{n_\ell^0\}_{\ell=1}^{m}\ \cup\ \{n_j^i\mid j=1,\dots,\bar m,\;S_j^i=1\}. \tag{18}
\label{eq:rev18}
\end{equation*}

The projection interval of a vertex set $\mathcal S$ along each axis $n\in\mathcal N^i$ is

\begin{equation*}
I_{\mathcal S}(n) =[\alpha_{\mathcal S}(n),\beta_{\mathcal S}(n)] :=\Big[\min_{q\in\mathcal S}\langle q,n\rangle,\ \max_{q\in\mathcal S}\langle q,n\rangle\Big]. \tag{19}
\label{eq:rev19}
\end{equation*}

Evaluating \eqref{eq:rev19} on the robot vertex set $\tilde V^0$ and the local obstacle vertex set $\tilde V^i$, we write

\begin{equation*}
\begin{aligned} I_0(n)&:=I_{\tilde V^0}(n)=[\alpha_0(n),\beta_0(n)],\\ I_i(n)&:=I_{\tilde V^i}(n)=[\alpha_i(n),\beta_i(n)]. \end{aligned}
\end{equation*}

Since both bodies are the convex hulls of these vertex sets, the intervals are the projections of the polygons themselves. The corresponding overlap along axis~$n$ is then defined as

\begin{equation*}
\delta^{i}(n):=\min\bigl\{\beta_i(n)-\alpha_0(n),\ \beta_0(n)-\alpha_i(n)\bigr\}. \tag{20}
\label{eq:rev20}
\end{equation*}

By the SAT, two convex polygons are disjoint if and only if their projections are disjoint along at least one edge normal of either polygon. A negative value of $\delta^{i}(n)$ identifies such a separating axis, whereas a positive value indicates overlapping projected intervals. Thus, in exact arithmetic, obstacle $i$ is separated from the footprint when

\begin{equation*}
\texttt{separated}^{i}:=\bigvee_{n\in\mathcal N^{i}}\big[\delta^{i}(n)<0\big]. \tag{21}
\label{eq:rev21}
\end{equation*}

When \eqref{eq:rev21} reports that the pair is not separated, the penetration depth between the robot footprint and obstacle $i$ is the smallest nonnegative axis-wise overlap,

\begin{equation*}
\delta(x,\mathcal E^i):=\min_{n\in\mathcal N^i}\max\big(\delta^{i}(n),0\big). \tag{22}
\label{eq:rev22}
\end{equation*}

Since $\mathcal N^i$ contains the unit edge normals of both convex polygons, this hard minimum gives the exact penetration depth $\operatorname{pen}(P^0(x),P^i)$, with zero at contact. All projections and reductions are evaluated in parallel with padded entries masked out.

\subsection{Signed Distance Synthesis and Differentiability}

Let $\operatorname{sd}^i(x):=\operatorname{sd}\bigl(P^{0}(x),P^{i}\bigr)$ denote the signed distance to obstacle $i$. Combining the nonnegative separation distance $d_{\mathrm{sep}}(x,\mathcal E^i)$ and penetration depth $\delta(x,\mathcal E^i)$ gives

\begin{equation*}
\operatorname{sd}^i(x)=
\begin{cases}
  d_{\mathrm{sep}}(x,\mathcal E^{i}), & \text{if }\texttt{separated}^{i}\\[4pt]
  -\,\delta(x,\mathcal E^{i}), & \text{otherwise}
\end{cases}, \tag{23}
\label{eq:rev23}
\end{equation*}

where $\texttt{separated}^{i}$ is the SAT-based separation status \eqref{eq:rev21}. Both branches yield zero at contact, ensuring continuity of $\operatorname{sd}^{i}$ across the switch. It is exact for the given polygonal geometry in either branch. The PSDF value $\phi(x,\mathcal E)$ is then obtained by taking the minimum of $\operatorname{sd}^{i}(x)$ over all obstacle components, as defined in \eqref{eq:rev6}.

The graph combines geometric operations, ReLU clamping, min/max reductions, and the selection in \eqref{eq:rev23}, yielding a piecewise differentiable PSDF. Automatic differentiation returns the state gradient $\nabla_x\phi(x,\mathcal E)$ where the active computation is differentiable. At points of nondifferentiability, such as min/max ties, ReLU kinks, and branch switches, the backward pass uses convex combinations of local derivative contributions or selected branch derivatives.

\section{Controller Design}
\label{controller_design}

\subsection{Linearized Safety Constraints within SQP-RTI}

Problem \eqref{eq:rev7} is solved with the real-time iteration (RTI) scheme \cite{diehl2005real}, which takes one sequential quadratic programming (SQP) step per sampling instant.

Let
\begin{equation*}
\bar X:=\{\bar x_{k+s}\}_{s=0}^{N}, \qquad \bar U:=\{\bar u_{k+s}\}_{s=0}^{N-1} \tag{24}
\label{eq:rev24}
\end{equation*}
denote the nominal state and input sequences at control step $k$. We obtain these sequences by shifting the state and input trajectories computed at the previous control step. The measured state $x_k$ is imposed through the initial-state constraint. The QP decision variables are the increments

\begin{equation*}
\Delta x_{k+s}:=x_{k+s}-\bar x_{k+s}, \qquad \Delta u_{k+s}:=u_{k+s}-\bar u_{k+s}. \tag{25}
\label{eq:rev25}
\end{equation*}

Following the standard RTI construction, the dynamics of \eqref{eq:rev7} are linearized about $(\bar X,\bar U)$, and its quadratic cost is expressed in terms of the state and input increments.

For each prediction stage $s=1,\ldots,N$, define $\phi_s:=\phi(\bar x_{k+s},\mathcal E_{k+s})$, $h_s:=\phi_s-d_{\min}$, and $J_{g,s}:=-\nabla_x\phi(\bar x_{k+s},\mathcal E_{k+s})$. Since the minimum clearance $d_{\min}$ is constant, the safety function $h$ has the same state gradient as the PSDF $\phi$, and a first-order expansion about $\bar x_{k+s}$ gives

\begin{equation*}
\begin{aligned} h(x_{k+s},\mathcal E_{k+s}) \approx h_s-J_{g,s}^{\top}\Delta x_{k+s}. \end{aligned} \tag{26}
\label{eq:rev26}
\end{equation*}

The following proposition provides a local tightening of this affine model to account for constraint-linearization error.

\begin{proposition}
Fix a prediction stage $s$ and hold $\mathcal E_{k+s}$ fixed. Let $r_{\max}:=\max_{v\in\tilde V^0}\lVert v\rVert_2$. Consider state increments $\Delta x_{k+s}$ whose pose components satisfy

\begin{equation*}
\lVert\Delta p_{k+s}\rVert_2\leq\rho_{p,s}, \qquad |\Delta\theta_{k+s}|\leq\rho_{\theta,s},
\end{equation*}

with positive radii. Assume that $\phi$ is differentiable at $\bar x_{k+s}$ and that, throughout this trust region, the robot footprint remains separated from all obstacles and the closest obstacle and vertex--edge pair remain the same. Define the tightening margin for constraint-linearization error as

\begin{equation*}
\begin{gathered} \gamma_s:=\frac12(r_{\max}+\phi_s)\rho_{\theta,s}^{2} +\rho_{p,s}\rho_{\theta,s}. \end{gathered} \tag{27}
\label{eq:rev27}
\end{equation*}

Under these assumptions, enforcing the bounds and the tightened linearized safety constraint

\begin{equation*}
J_{g,s}^{\top}\Delta x_{k+s}\leq h_s-\gamma_s \tag{28}
\label{eq:rev28}
\end{equation*}

guarantees that the updated predicted state satisfies the nonlinear safety constraint $\phi(\bar x_{k+s}+\Delta x_{k+s},\mathcal E_{k+s})\geq d_{\min}$.

\end{proposition}

\begin{IEEEproof}
\IEEEQEDoff
Let $e_s$ denote the displacement from the first-order approximation of the vertex position to its exact position, expressed in the frame where the active edge is fixed.

For the first pool \eqref{eq:rev14}, the obstacle edge is fixed in $\mathcal W$, while the robot vertex is $p+R(\theta)\tilde v_\ell^0$. Translation is affine, so only rotation contributes to the remainder. Since $\|R''(\theta)\|_2=1$, the remainder satisfies $\|e_s\|_2\leq\frac12r_{\max}\rho_{\theta,s}^2$.

For the second pool \eqref{eq:rev15}, the robot edge is fixed in $\mathcal L$, while the obstacle vertex is $R(\theta)^\top(A_j^i-p)$. The remainder contains rotation and rotation--translation coupling terms, giving

\begin{equation*}
\|e_s\|_2 \leq \frac12\|A_j^i-\bar p_{k+s}\|_2\rho_{\theta,s}^2 +\rho_{p,s}\rho_{\theta,s}.
\end{equation*}

The nominal closest point on the robot edge is at most $r_{\max}$ from the nominal robot position, so the triangle inequality gives $\|A_j^i-\bar p_{k+s}\|_2\leq r_{\max}+\phi_s$. Thus, $\|e_s\|_2\leq\gamma_s$ in either pool.

The point-to-segment distance is convex in the point coordinates, with unit gradient $n_s$ at the nominal point. Its supporting-tangent inequality, the chain rule, and \eqref{eq:rev28} give

\begin{equation*}
\begin{aligned}
  \phi(\bar x_{k+s}+\Delta x_{k+s},\mathcal E_{k+s})
    &\geq \phi_s-J_{g,s}^{\top}\Delta x_{k+s}+n_s^{\top}e_s\\
    &\geq \phi_s-J_{g,s}^{\top}\Delta x_{k+s}-\gamma_s
      \geq d_{\min}.
\end{aligned}
\end{equation*}

\end{IEEEproof}

The guarantee is local under the fixed-feature assumption and does not cover feature switching. In the implementation, we use fixed radii $\rho_{p,s}=v_{\max}\Delta t$ and $\rho_{\theta,s}=\omega_{\max}\Delta t$, where $v_{\max}$ and $\omega_{\max}$ denote the maximum allowable linear and angular speeds, respectively, and $\Delta t$ is the sampling period. The corresponding SQP pose-increment bounds are enforced as $\|\Delta p_{k+s}\|_{\infty}\leq\frac{\rho_{p,s}}{\sqrt{2}}$ and $|\Delta\theta_{k+s}|\leq\rho_{\theta,s}$ for $s=1,\ldots,N$, preserving the QP structure. With the tightening margin $\gamma_s$ fixed during each solve, \eqref{eq:rev28} contributes one linearized safety constraint per prediction stage. Obstacle geometry enters the QP through the stage-wise parameter vectors $\{[h_s-\gamma_s;\,J_{g,s}]\}_{s=1}^{N}$, so obstacle and edge counts do not increase the number of decision variables or constraint rows.

\subsection{Real-Time Computational Pipeline}

Each control cycle consists of one batched PSDF value and gradient evaluation on the GPU and one QP solve on the CPU. Given the nominal state and input sequences $(\bar X,\bar U)$, the GPU receives the $N$ nominal states $\{\bar x_{k+s}\}_{s=1}^{N}$ as a single batch, evaluates the corresponding PSDF values and state gradients, and forms the coefficients $(h_s-\gamma_s,J_{g,s})$ of \eqref{eq:rev28}. These coefficients are transferred to the CPU, which assembles and solves the QP subproblem once. The edge set is frozen over the prediction horizon, $\mathcal E_{k+s}\equiv\mathcal E_{k}$, so that one encoding of the environment serves the entire batch. A time-varying environment could be accommodated by batching $\{(\bar x_{k+s},\mathcal E_{k+s})\}_{s=1}^{N}$ instead, which enlarges the padded tensors of \eqref{eq:rev10} but leaves the number of QP rows and their sparsity pattern unchanged. Algorithm~1 summarizes the control cycle for $\mathcal E := \mathcal E_k$. Here, $\Phi := [\phi_1,\ldots,\phi_N]^\top$ denotes the vector of PSDF values over the prediction stages, with $\Phi_s=\phi_s$.

The PyTorch PSDF module is connected to the acados solver through CasADi and the real-time L4CasADi interface \cite{salzmann2024learning,salzmann2023real}, using a first-order local Taylor surrogate whose coefficients are updated online. The geometric computation graph is evaluated in PyTorch, outside the solver. Transformations, projections, and distance computations are implemented using batched tensor operations. Padded entries are excluded from the minimum and maximum reductions (\texttt{torch.amin}, \texttt{torch.amax}), and the selection in \eqref{eq:rev23} is implemented with elementwise tensor operations. The resulting graph is branch-free. One batched forward pass and one backward pass compute PSDF values and state gradients for all $N$ stages.

\begin{algorithm}[t]
\caption{PSDF-MPC with separated CPU/GPU execution}
\label{alg:psdf_mpc_pipeline}
\small
\algrenewcommand{\algorithmicrequire}{\textbf{Inputs:}}
\algrenewcommand{\algorithmicensure}{\textbf{Outputs:}}
\begin{algorithmic}[1]
\setlength{\itemsep}{1pt}
\Require $x_k$; $(\bar X,\bar U)$; $\mathcal E$; $N$; $d_{\min}$.
\Ensure $u_k$; $(\bar X,\bar U)$ for step $k+1$.
\Statex \rule{\linewidth}{0.4pt}
\Statex \textbf{GPU: batched PSDF evaluation over the horizon}
\Statex $(s=1,\ldots,N)$
\If{$\mathcal E$ is new or changed}
\State Upload $\mathcal E$ to GPU
\State $(A,B,S)\gets\mathrm{EncodeEdges}(\mathcal E)$
\EndIf
\State Upload $\{\bar x_{k+s}\}_{s=1}^{N}$ to GPU
\State $\Phi_s\gets\phi(\bar x_{k+s},\mathcal E)$ \Comment{batched forward}
\State $J_{g,s}\gets-\nabla_x\phi(\bar x_{k+s},\mathcal E)$ \Comment{batched backward}
\State $h_s\gets\Phi_s-d_{\min}$
\State Compute the tightening margin $\gamma_s$ using \eqref{eq:rev27}
\State $\boldsymbol p\gets\{[h_s-\gamma_s;\,J_{g,s}]\}_{s=1}^{N}$
\State Transfer $\boldsymbol p$ to CPU
\Statex \rule{\linewidth}{0.4pt}
\Statex \textbf{CPU: one SQP-RTI QP solve and input application}
\State Impose the initial-state constraint; enforce pose-increment bounds for $s=1,\ldots,N$.
\State Inject $\boldsymbol p$ into the QP subproblem
\Statex \makebox[\linewidth][c]{$J_{g,s}^{\top}\Delta x_{k+s}\leq h_s-\gamma_s,\quad s=1,\ldots,N$}
\State $(\Delta X^\star,\Delta U^\star)\gets\mathrm{SolveRTIQP}(\bar X,\bar U)$
\State Apply $u_k\gets\bar u_k+\Delta u_k^\star$
\State $(\bar X,\bar U)\gets\mathrm{Shift}(\bar X+\Delta X^\star,\bar U+\Delta U^\star)$
\end{algorithmic}
\end{algorithm}

\raggedbottom
\setlength{\textfloatsep}{10pt plus 2pt minus 2pt}
\setlength{\dbltextfloatsep}{10pt plus 2pt minus 2pt}
\section{EXPERIMENTS}
\label{experiments}

\subsection{Experimental Setup}

The experiments evaluate the computational scalability of PSDF and assess the control-update time and navigation performance of PSDF-MPC as free space narrows. Further experiments examine real-time onboard execution and applicability to Ackermann parking while retaining the same PSDF-based safety constraints. The query microbenchmark, the planar maze and Gazebo navigation experiments, and the CARLA parking study were run on a desktop workstation with an Intel Core i9-13900KF CPU and an NVIDIA RTX 4090 GPU. Only the onboard experiment was run on the robot itself. The Scout Mini platform carried an NVIDIA Jetson Xavier, and obstacle geometry was obtained from a Velodyne VLP-16 LiDAR. In all navigation experiments, PSDF-MPC used a prediction horizon of $N=15$, a sampling period of $\Delta t = 0.1$ s, and a minimum clearance of $d_{\min} = 0.01$ m. Per-step computation time is defined as the wall-clock duration from receiving a state measurement to computing the corresponding control command; this timing definition is used in the planar and Gazebo navigation experiments and the onboard deployment.

For each environment in the planar and Gazebo experiments, all methods were evaluated over 100 trials using the same initial poses, generated by adding independent zero-mean Gaussian perturbations to the nominal pose ($\sigma_x=\sigma_y=0.05\,\mathrm{m}$, $\sigma_\theta=0.1\,\mathrm{rad}$). Success rate is the percentage of trials in which the robot reaches the goal without collision or solver failure. We report the mean and 95th percentile (P95) of computation time over all recorded control updates. Gazebo navigation time is averaged over successful trials only.

\begin{table}[t]
\centering
\caption{Total latency of value and state-gradient evaluation for collision avoidance.}
\label{tab:sdf_ctime}
\footnotesize
\setlength{\tabcolsep}{5pt}
\renewcommand{\arraystretch}{1.15}
\begin{tabular*}{\columnwidth}{@{\extracolsep{\fill}}lcccc@{}}
\toprule
\multirow{2}{*}{\textbf{Method}} &
\multirow{2}{*}{\textbf{$M$}} &
\multicolumn{3}{c}{\makecell{\textbf{Total evaluation latency [ms]}\\Mean (Std)}} \\
\cmidrule(lr){3-5}
& & \textbf{$B=1$} & \textbf{$B=15$} & \textbf{$B=100$} \\
\midrule
\multirow{3}{*}{PSDF}
&   1 & 1.981\,(0.074) & 2.154\,(0.039) & 2.212\,(0.131) \\
&  10 & 2.152\,(0.055) & 2.213\,(0.104) & 2.250\,(0.182) \\
& 100 & 2.221\,(0.113) & 2.222\,(0.038) & 2.207\,(0.094) \\
\midrule
\multirow{3}{*}{GJK{+}EPA}
&   1 & 0.014\,(0.012) & 0.160\,(0.009) & 1.086\,(0.157) \\
&  10 & 0.060\,(0.008) & 0.886\,(0.017) & 5.845\,(0.024) \\
& 100 & 0.502\,(0.017) & 8.012\,(0.045) & 54.081\,(0.721) \\
\midrule
\multirow{3}{*}{NPField}
&   1 & 3.500\,(0.069) & 3.413\,(0.056) & 3.455\,(0.075) \\
&  10 & 3.442\,(0.074) & 3.458\,(0.074) & 3.508\,(0.142) \\
& 100 & 3.453\,(0.092) & 3.441\,(0.042) & 3.454\,(0.060) \\
\bottomrule
\end{tabular*}
\end{table}

\begin{figure}[t]
  \centering
  \subfloat[\label{fig:fig2-1-1}]{%
    \includegraphics[width=0.33\columnwidth]{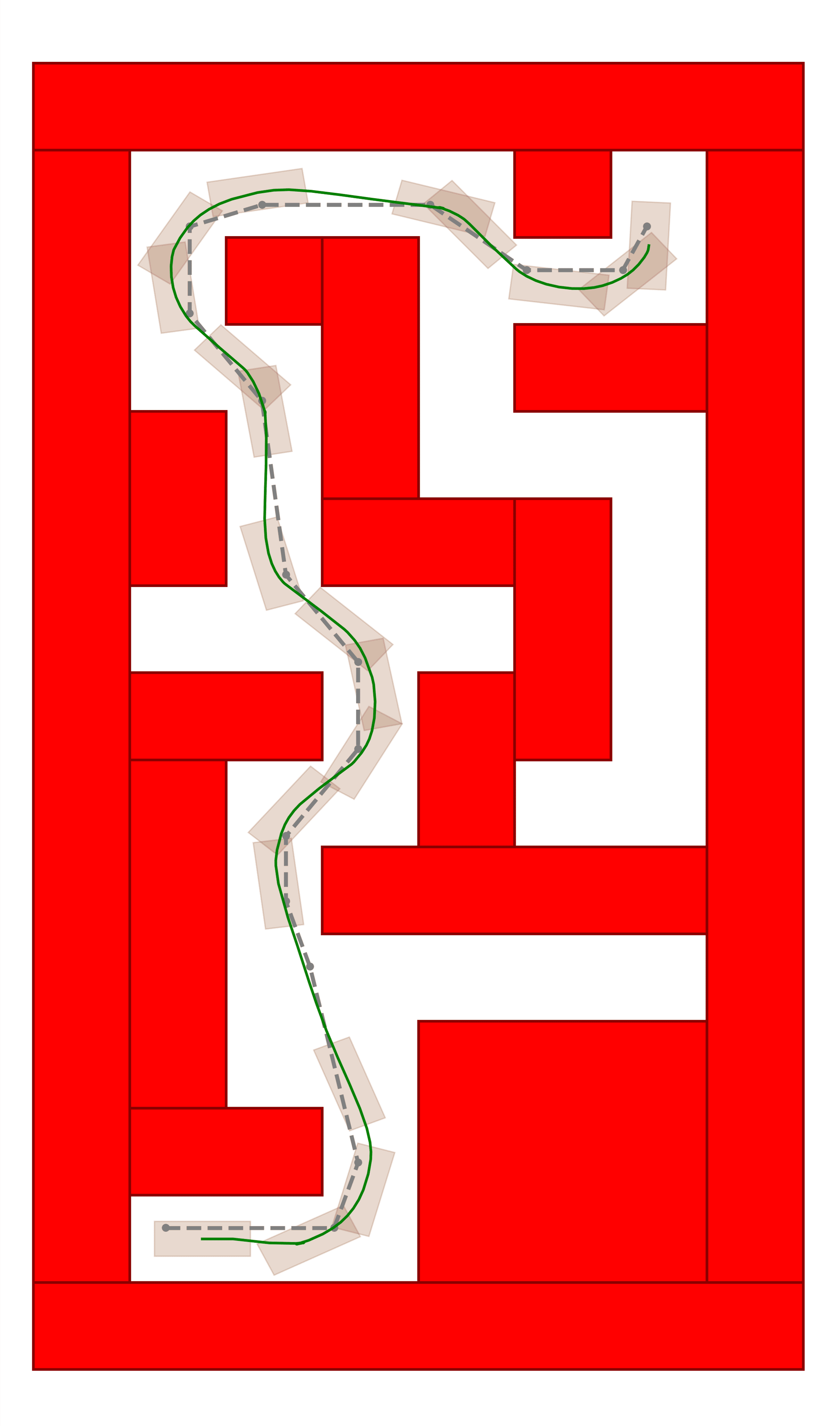}}\hfill
  \subfloat[\label{fig:fig2-2-1}]{%
    \includegraphics[width=0.33\columnwidth]{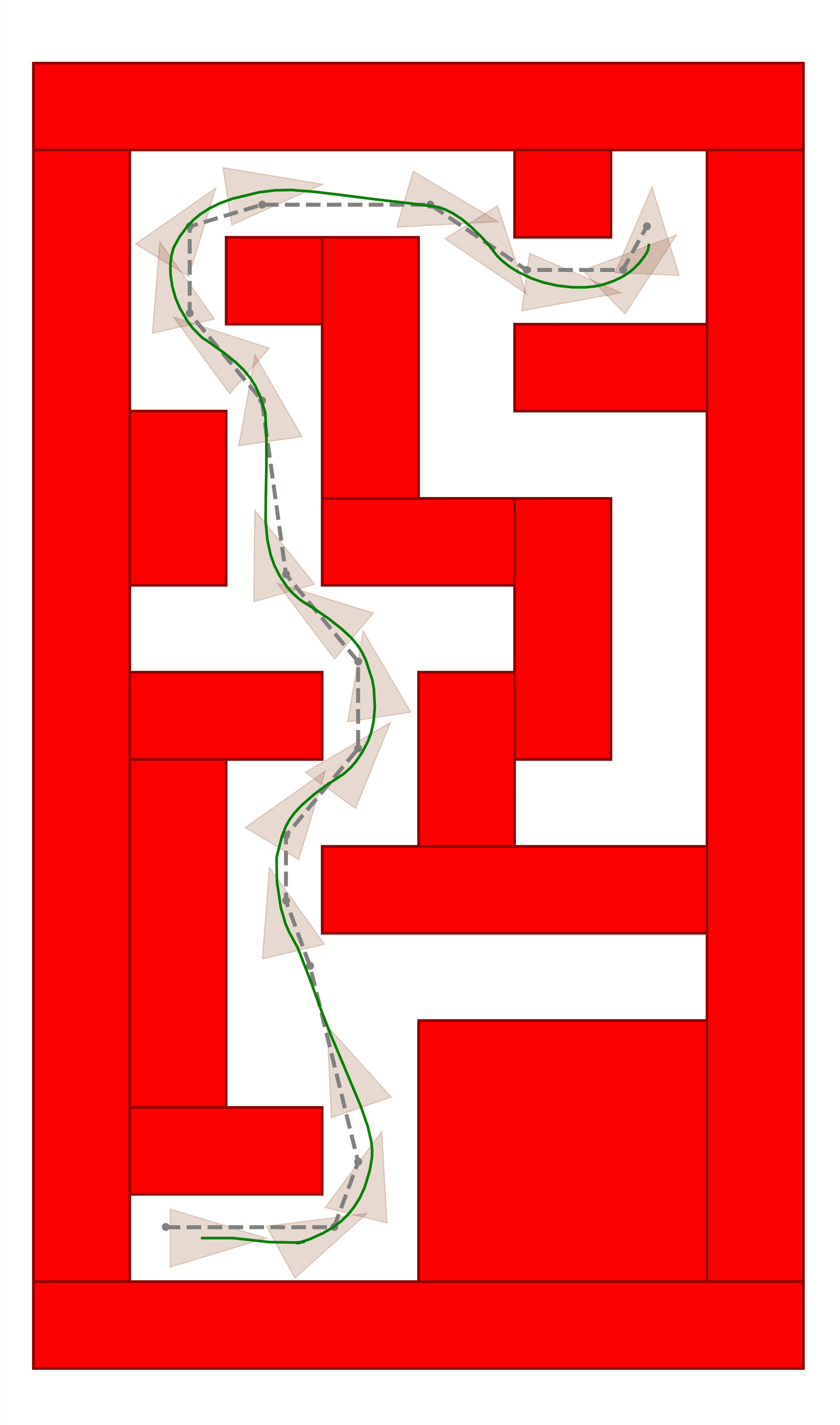}}\hfill
  \subfloat[\label{fig:fig2-3-1}]{%
    \includegraphics[width=0.33\columnwidth]{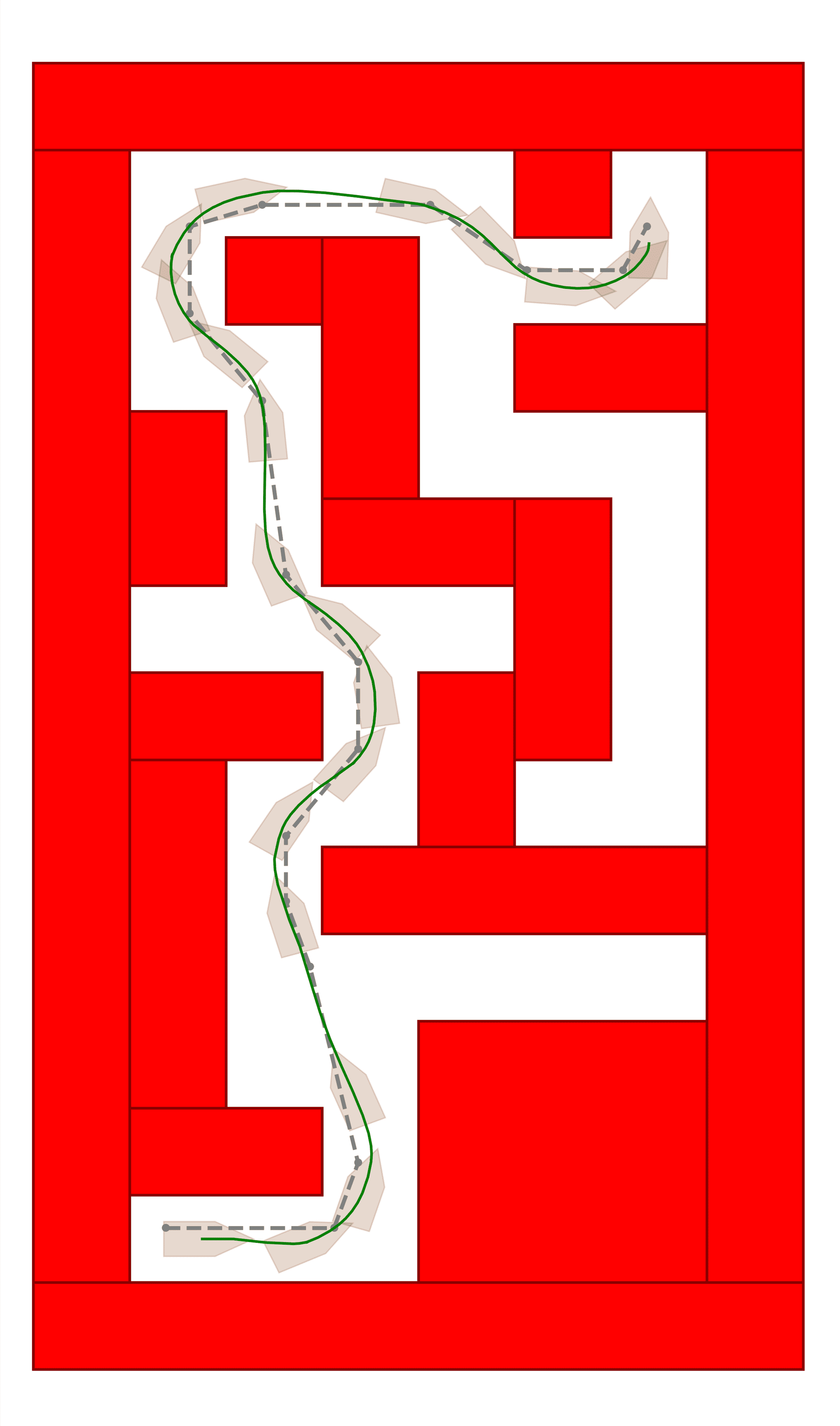}}

  \vspace{0.6em}

  \subfloat[\label{fig:fig2-1-2}]{%
    \includegraphics[width=0.33\columnwidth]{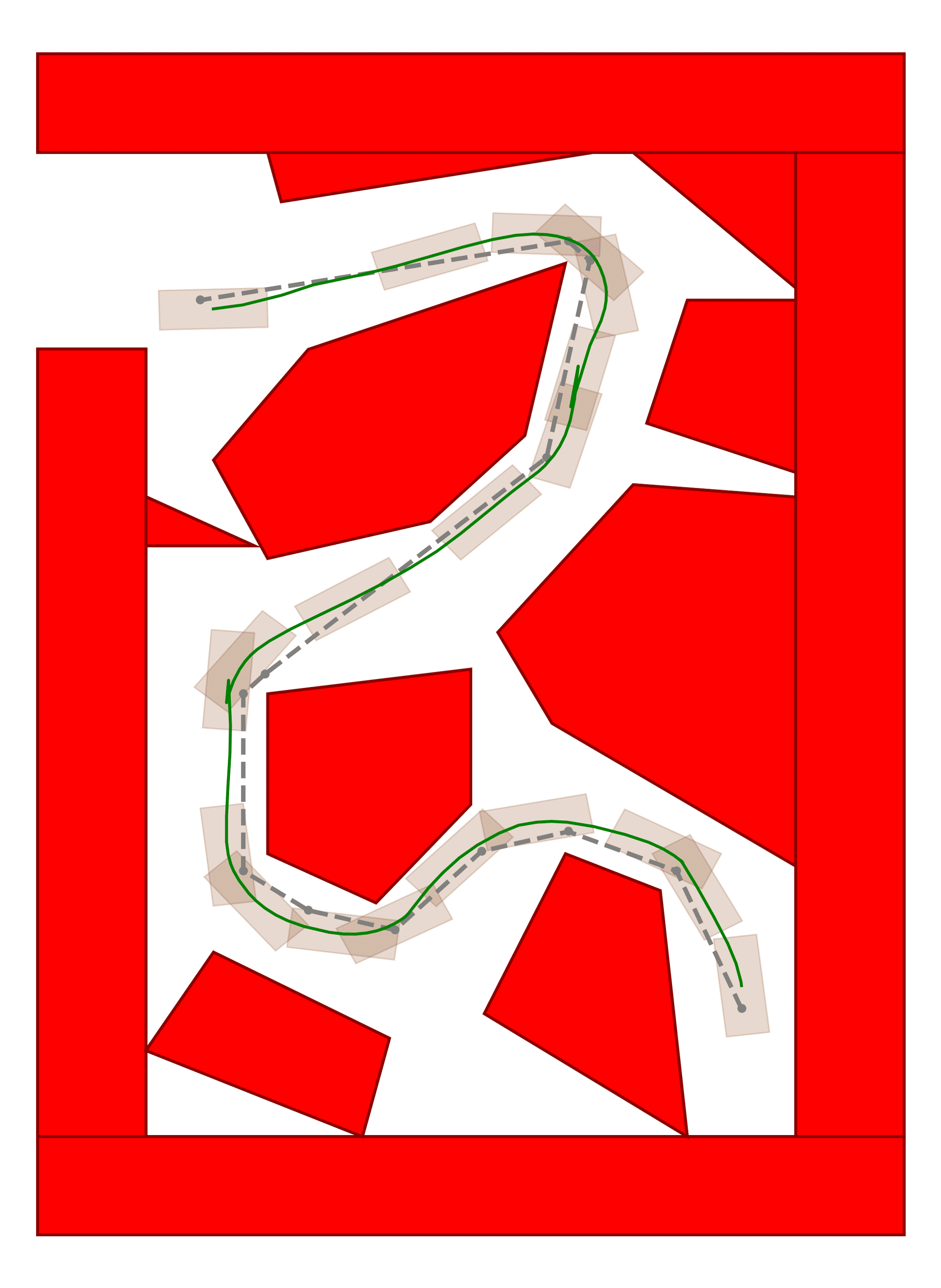}}\hfill
  \subfloat[\label{fig:fig2-2-2}]{%
    \includegraphics[width=0.33\columnwidth]{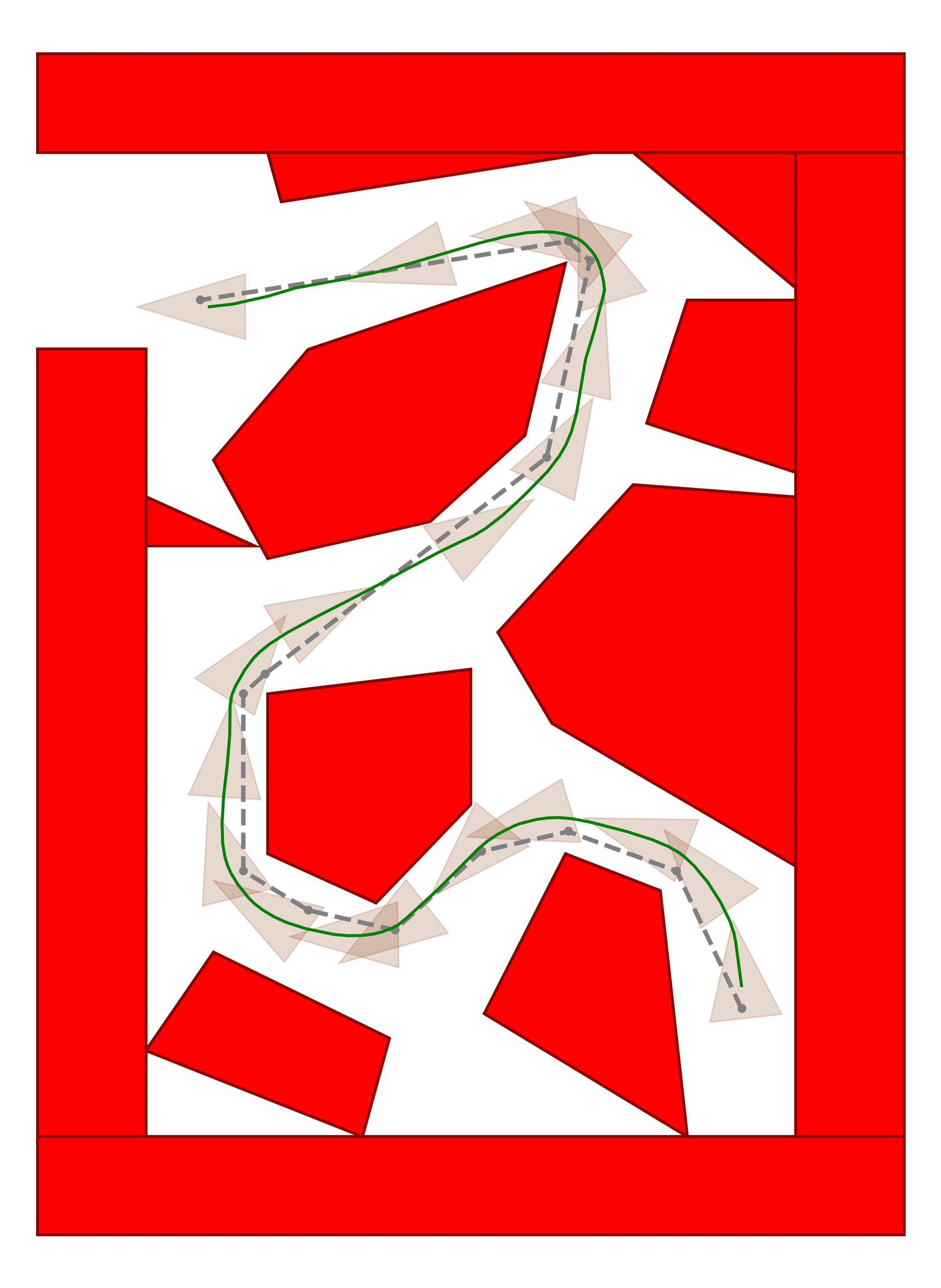}}\hfill
  \subfloat[\label{fig:fig2-3-2}]{%
    \includegraphics[width=0.33\columnwidth]{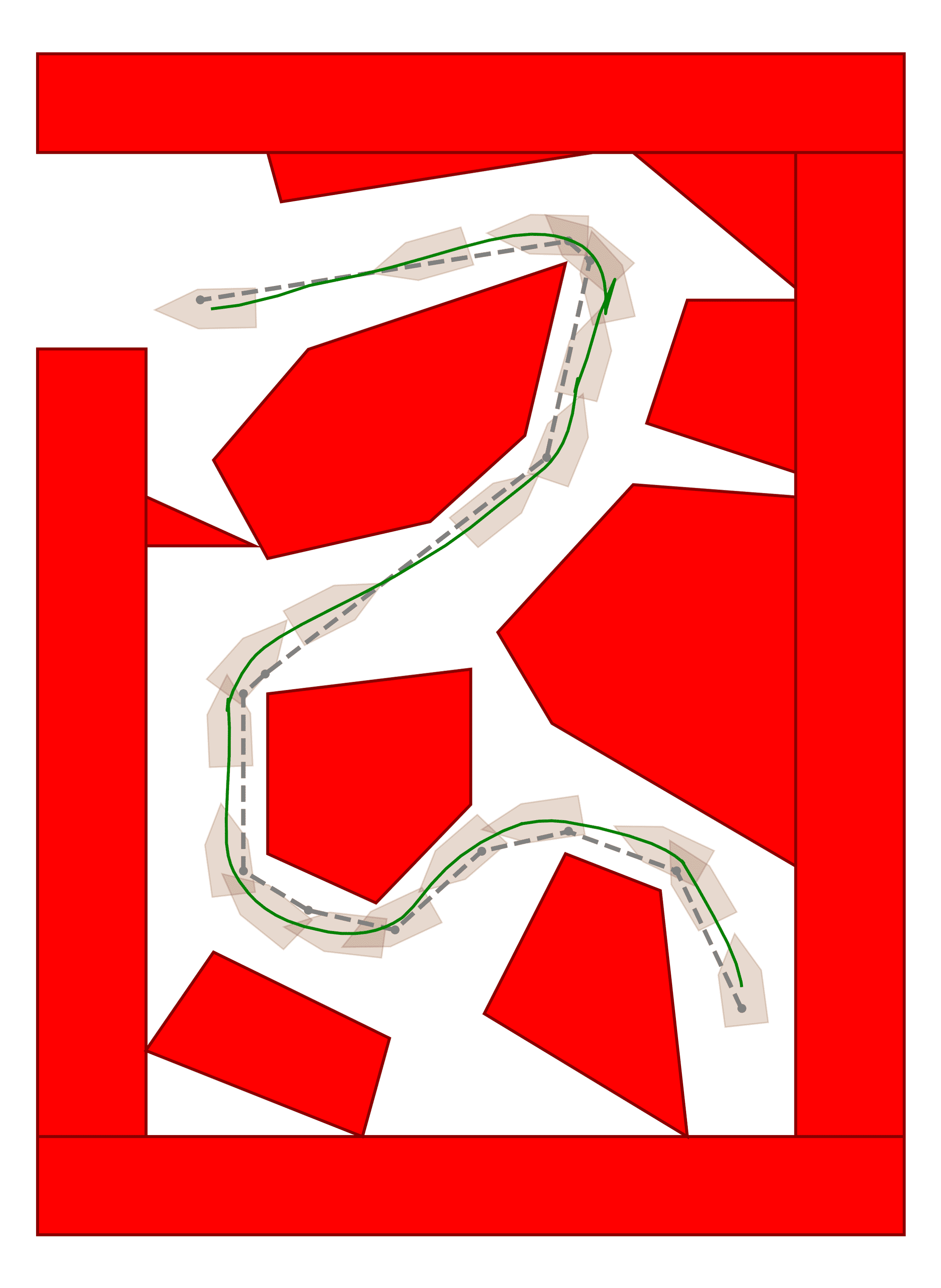}}

  \caption{Closed-loop trajectories generated by PSDF-MPC in the 2D Maze environments. (a)--(c) show the Maze environment, and (d)--(f) show the Oblique Maze.
  The robot footprint is modeled as a rectangle, triangle, and pentagon (left to right).
  Dashed lines denote the reference trajectories, and green lines show the actual trajectories.}
  \label{fig:maze_traj}
\end{figure}

\begin{table}[t]
  \centering
  \caption{Computation time in 2D maze environments with a rectangular robot footprint.}
  \label{tab:opt_time}
  \scriptsize
  \setlength{\tabcolsep}{5pt}
  \renewcommand{\arraystretch}{1.15}
  \begin{tabular}{@{}llc@{}}
    \toprule
    \textbf{Environment} & \textbf{Method} &
    \makecell{\textbf{Computation time [ms]}\\Mean / P95} \\
    \midrule
    \multirow{3}{*}{Maze}
      & PSDF-MPC & \textbf{19.8 / 25.6} \\
      & OBCA     & 230.5 / 502.3 \\
      & DCBF     & 193.0 / 437.6 \\
    \midrule
    \multirow{3}{*}{Oblique Maze}
      & PSDF-MPC & \textbf{16.9 / 22.4} \\
      & OBCA     & 166.7 / 388.4 \\
      & DCBF     & 140.5 / 341.0 \\
    \bottomrule
  \end{tabular}
\end{table}

\subsection{Latency of Value and Gradient Evaluation for Collision Avoidance}

This benchmark evaluates the computational cost of obtaining scalar values and their state gradients for collision avoidance. The reported latency is the total time required to compute the value and state gradient of each collision representation. For NPField, it additionally includes the cost of updating the environment representation through map and robot-footprint encoding.

PSDF was compared against GJK+EPA \cite{gilbert1988fast,van2001proximity} and NPField \cite{alhaddad2024neural} on randomly generated obstacle sets and query poses, sweeping obstacle count $M\in\{1,10,100\}$ and batch size $B\in\{1,15,100\}$, where $B$ is the number of poses evaluated per call, with the mean and standard deviation of total evaluation latency reported in Table~\ref{tab:sdf_ctime}. In this benchmark, the padded obstacle capacity $\bar M$ was increased with $M$. The methods differ in how gradients are obtained: PSDF and NPField are GPU modules differentiated by automatic differentiation, whereas GJK+EPA runs on the CPU and constructs pose gradients from witness points and normals. NPField is included as a learned collision-field reference, for which the evaluated quantities are the repulsive potential and its pose gradient.

PSDF latency varied only weakly with $M$ and $B$ over the tested range, consistent with batched GPU evaluation. NPField latency also varied little with $M$ and $B$ but remained higher because images of the map and robot footprint must be encoded whenever the environment changes. This encoding takes roughly the same time regardless of scene complexity. GJK+EPA remained faster for small workloads, whereas the advantage of PSDF became evident for larger batched workloads. With the batch size used in the PSDF-MPC implementation ($B=N=15$) and $M=100$, PSDF required 2.222 ms, compared with 8.012 ms for GJK+EPA. Thus, PSDF primarily improves the efficiency of batched evaluation over the prediction horizon in scenes with many obstacles, rather than providing consistently faster individual robot--obstacle queries.

\begin{figure}[!t]
  \centering
  \includegraphics[width=\columnwidth]{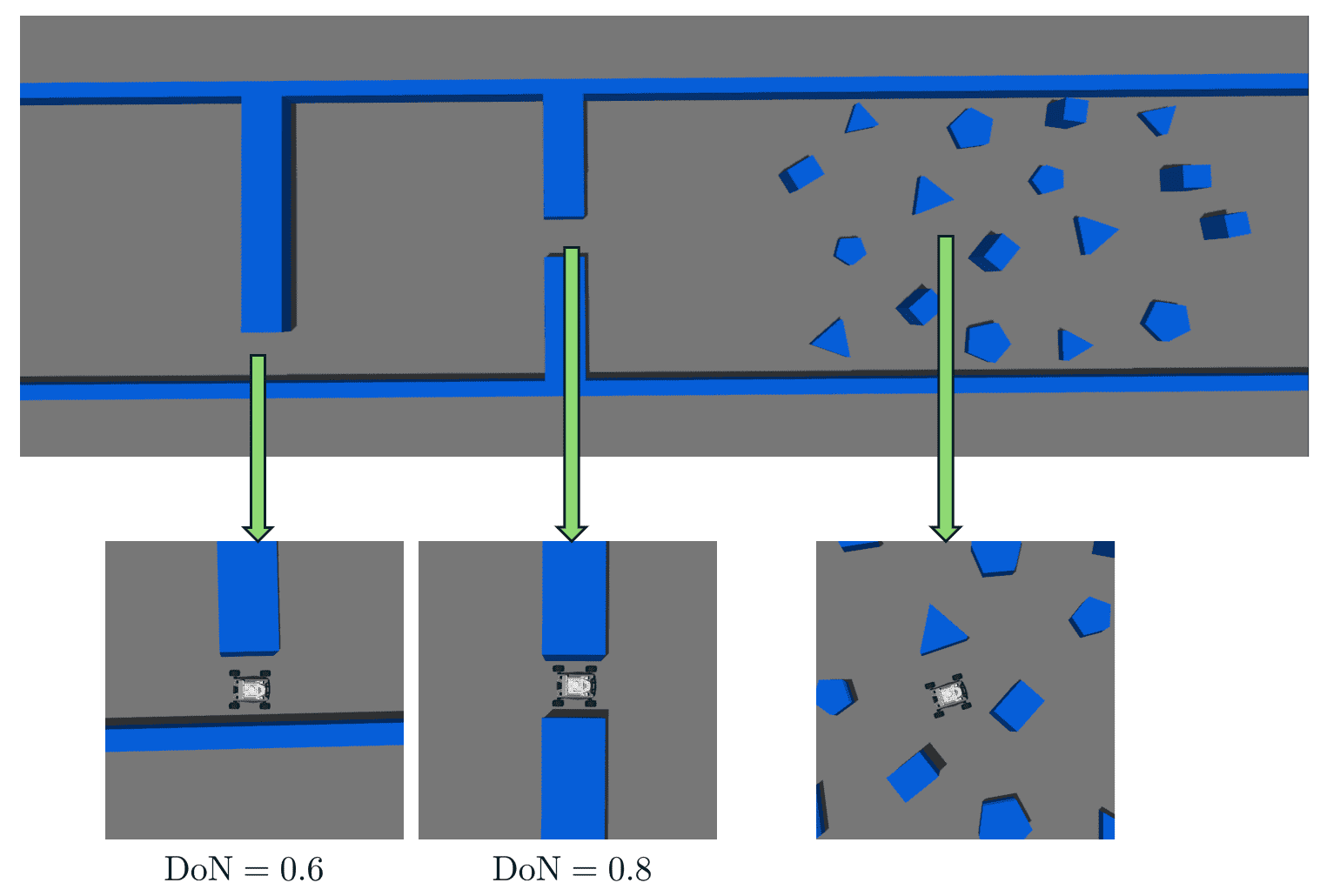}

  \caption{Gazebo test environment comprising \emph{Broad Corridor},
  \emph{Narrow Corridor}, and \emph{Cluttered Obstacle Field}.}
  \label{fig:gazebo_traj}
\end{figure}

\begin{table*}[t]
\centering
\setlength{\tabcolsep}{4pt}
\renewcommand{\arraystretch}{1.15}
\caption{Performance comparison of PSDF\mbox{-}MPC (ours), TEB, and RDA in the Gazebo navigation benchmark.}
\label{tab:psdf_teb_rda}
\resizebox{\textwidth}{!}{%
\begin{tabular}{@{}lccc|ccc|ccc@{}}
\toprule
\multirow{2}{*}{\textbf{Scenario}} &
\multicolumn{3}{c|}{\textbf{Success Rate [\%]}} &
\multicolumn{3}{c|}{\textbf{Mean Navigation Time [s]}} &
\multicolumn{3}{c}{\makecell{\textbf{Computation time [ms]}\\Mean / P95}} \\
\cmidrule(lr){2-4}\cmidrule(lr){5-7}\cmidrule(lr){8-10}
& \makecell{PSDF-MPC\\(ours)} & TEB & RDA
& \makecell{PSDF-MPC\\(ours)} & TEB & RDA
& \makecell{PSDF-MPC\\(ours)} & TEB & RDA \\
\midrule
Broad Corridor &
100 & 82 & 100 &
17.29 & 18.52 & 17.40 &
8.9 / 10.7 & 4.4 / 7.6 & 191.1 / 288.3 \\
Narrow Corridor &
85 & 0 & 71 &
12.16 & --- & 13.91 &
10.0 / 11.0 & --- & 247.4 / 365.9 \\
Cluttered Obstacle Field &
100 & 91 & 100 &
25.49 & 38.46 & 31.19 &
8.5 / 11.4 & 6.0 / 9.7 & 264.9 / 333.5 \\
\bottomrule
\end{tabular}%
}
\end{table*}

\begin{figure}[!t]
  \centering
  \begin{minipage}[t]{0.61\linewidth}
    \vspace{0pt}
    \centering
    \subfloat[\label{fig:fig4-a}]{%
      \includegraphics[width=\linewidth]{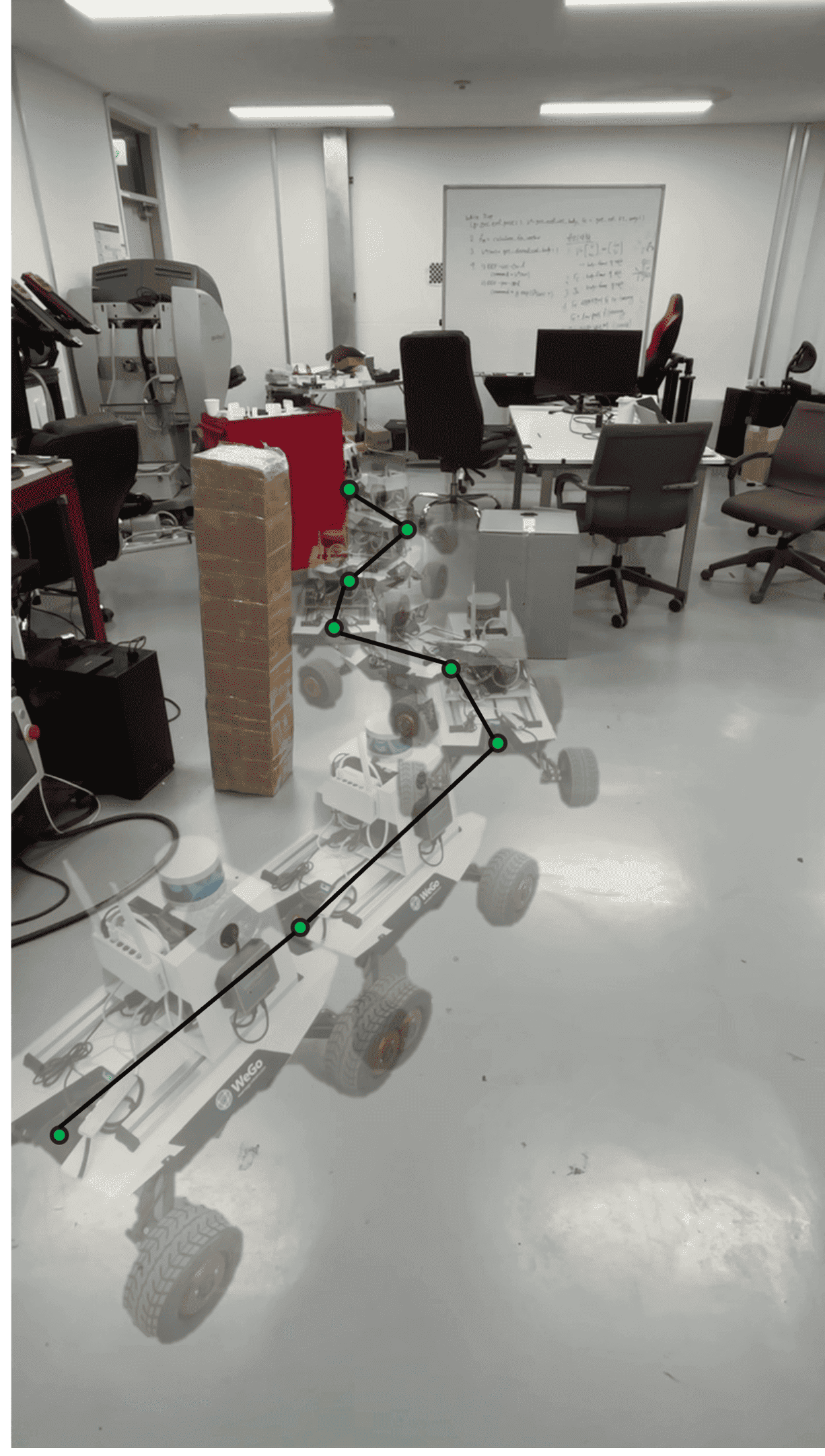}}
  \end{minipage}\hfill
  \begin{minipage}[t]{0.37\linewidth}
    \vspace{0pt}
    \centering
    \subfloat[\label{fig:fig4-b}]{%
      \includegraphics[width=\linewidth]{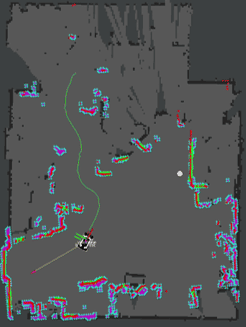}}

    \vspace{3pt}

    \subfloat[\label{fig:fig4-c}]{%
      \includegraphics[width=\linewidth]{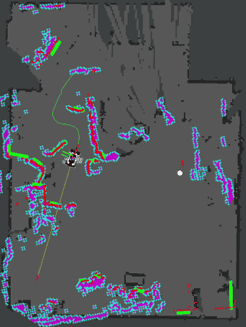}}
  \end{minipage}

  \caption{Real robot deployment: trajectories and top-down views.
    (a) Real-world driving environment and the trajectory executed by the robot.
    (b), (c) Top-down views of the perceived environment; the green regions
    correspond to safety-inflated rectangles constructed from boundaries fitted
    to the projected point cloud, which are used as inputs to PSDF.}
  \label{fig:real_traj}
\end{figure}

\subsection{Closed-Loop Navigation}

Two experiments evaluate the real-time feasibility and practical navigation performance of PSDF-MPC. Planar simulations in Python compare the computational efficiency of PSDF-MPC with that of OBCA \cite{zhang2020optimization} and DCBF \cite{thirugnanam2022safety}, both of which impose geometry-exact collision-avoidance constraints. All three methods use the same robot model, prediction horizon, and main cost terms.

The planar simulations use two obstacle-dense environments, Maze and Oblique Maze. PSDF-MPC is tested with rectangular, triangular, and pentagonal footprints by changing only the local vertex set $\tilde V^0$, while keeping the controller and PSDF implementation unchanged. Fig.~\ref{fig:maze_traj} shows representative closed-loop trajectories, and Table~\ref{tab:opt_time} reports the mean and P95 per-step computation times for the rectangular footprint.

In both maze environments, all controllers completed every trial successfully. PSDF-MPC achieved mean computation times of 16.9--19.8 ms and P95 values of 22.4--25.6 ms, yielding an approximately 8.3--11.6-fold speedup over OBCA and DCBF in terms of mean computation time. Both the mean and P95 values remained below the 100-ms sampling period, whereas the mean computation times of both baselines exceeded this period in both environments. These results demonstrate the real-time feasibility of PSDF-MPC in cluttered environments with complex polygonal geometry.

The Gazebo mobile robot simulation evaluates navigation in obstacle-rich environments using a system that integrates perception, localization, and control, with TEB \cite{rosmann2017integrated} and RDA \cite{han2023rda} as baselines. The Gazebo benchmark comprises three scenarios with different passage widths and obstacle densities: Broad Corridor, Narrow Corridor, and Cluttered Obstacle Field. Their tightness is characterized by the degree of narrowness (DoN), the ratio of the robot width to the minimum passage width \cite{han2025neupan}. At each control cycle, the 3D LiDAR point cloud is projected onto the ground plane, and obstacle boundaries are extracted by line fitting. Each fitted boundary is represented by a safety-inflated rectangle, whose edges form the polygonal edge set $\mathcal E$ supplied to PSDF. This inflation provides a safety buffer around the perceived obstacles. Fig.~\ref{fig:gazebo_traj} shows the three test environments, and Table~\ref{tab:psdf_teb_rda} summarizes navigation performance.

In the Gazebo benchmark, PSDF-MPC achieved success rates of 100\%, 85\%, and 100\% in the Broad Corridor, Narrow Corridor, and Cluttered Obstacle Field, respectively, and the shortest mean navigation time among successful trials in each scenario. TEB did not complete any Narrow Corridor trial. In the Cluttered Obstacle Field, TEB achieved 91\% success but required 38.46 s, compared with 25.49 s for PSDF-MPC. Although its mean and P95 computation times were only 6.0 and 9.7 ms, TEB exhibited repeated stops and local trajectory adjustments, interrupting reference tracking while attempting to maintain the desired clearance. This behavior is consistent with competition between penalty-based obstacle avoidance and other motion objectives under the tested configuration, rather than excessive computation time.

RDA achieved 100\% success in the Broad Corridor and Cluttered Obstacle Field, but 71\% in the Narrow Corridor. Its mean computation time increased from 191.1 ms in the Broad Corridor to 264.9 ms in the Cluttered Obstacle Field, consistent with the obstacle-dependent workload of its duality-based formulation. In the latter scenario, its P95 computation time reached 333.5 ms. These latencies exceeded the 100-ms sampling period, suggesting that delayed control updates contributed to the observed repeated stops and corrective maneuvers and the longer navigation time of 31.19 s.

In contrast, PSDF-MPC maintained mean computation times of 8.5--10.0 ms and P95 values of 10.7--11.4 ms across all three scenarios, supporting real-time feasibility without the increase in update latency observed for RDA. However, in all 15 unsuccessful PSDF-MPC trials in the Narrow Corridor, the robot stalled and could not make further progress toward the goal. These failures indicate that finding feasible trajectories remains challenging for PSDF-MPC when free space is severely restricted.

\subsection{Onboard Deployment and Ackermann Parking}

\begin{figure*}[!t]
  \centering
  \begin{minipage}[t]{0.48\textwidth}
    \vspace{0pt}
    \centering
    \subfloat[\label{fig:carla_traj-a}]{%
      \includegraphics[width=\linewidth]{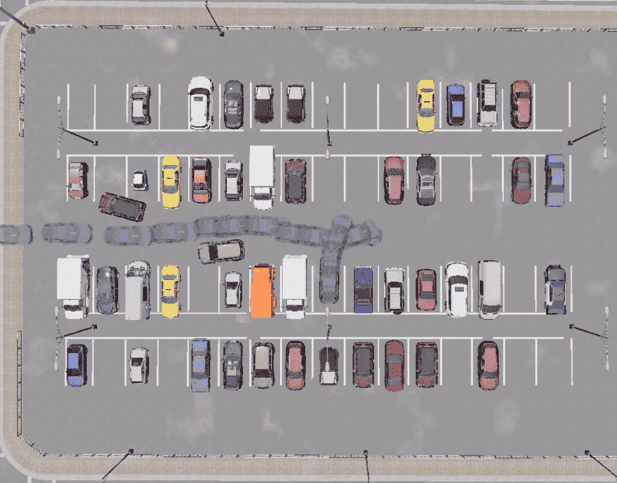}}
  \end{minipage}\hfill
  \begin{minipage}[t]{0.50\textwidth}
    \vspace{0pt}
    \centering
    \subfloat[\label{fig:carla_traj-b}]{%
      \includegraphics[width=\linewidth]{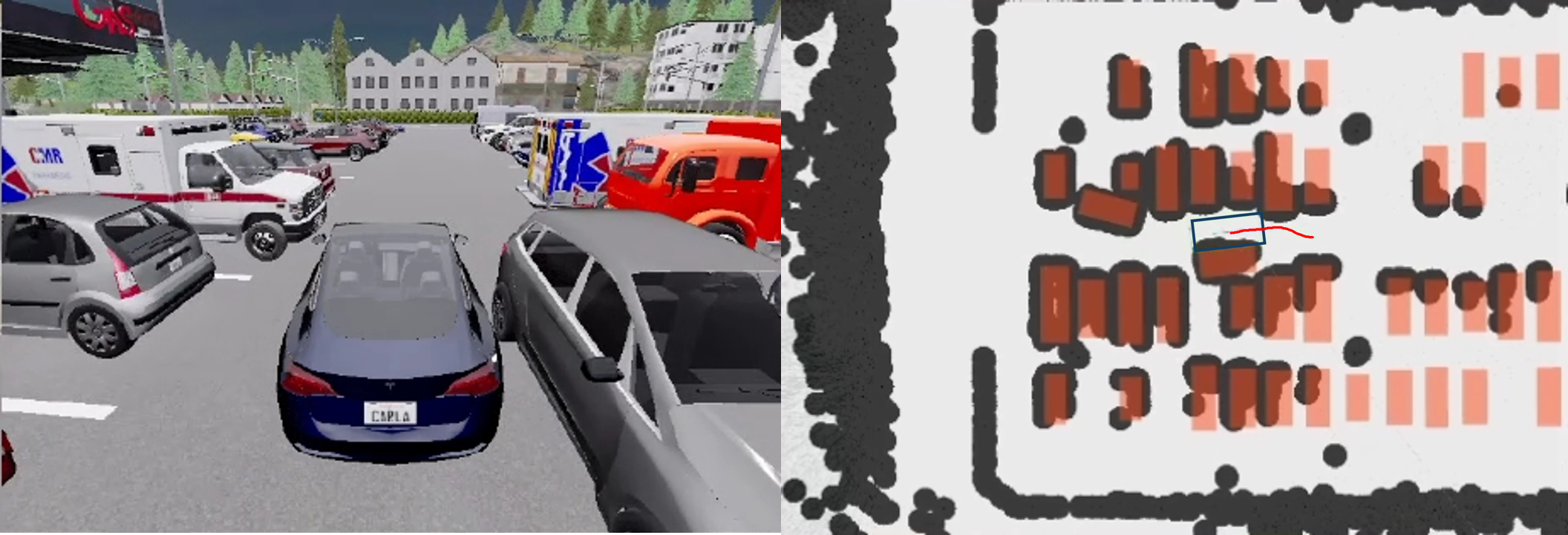}}

    \vspace{3pt}

    \subfloat[\label{fig:carla_traj-c}]{%
      \includegraphics[width=\linewidth]{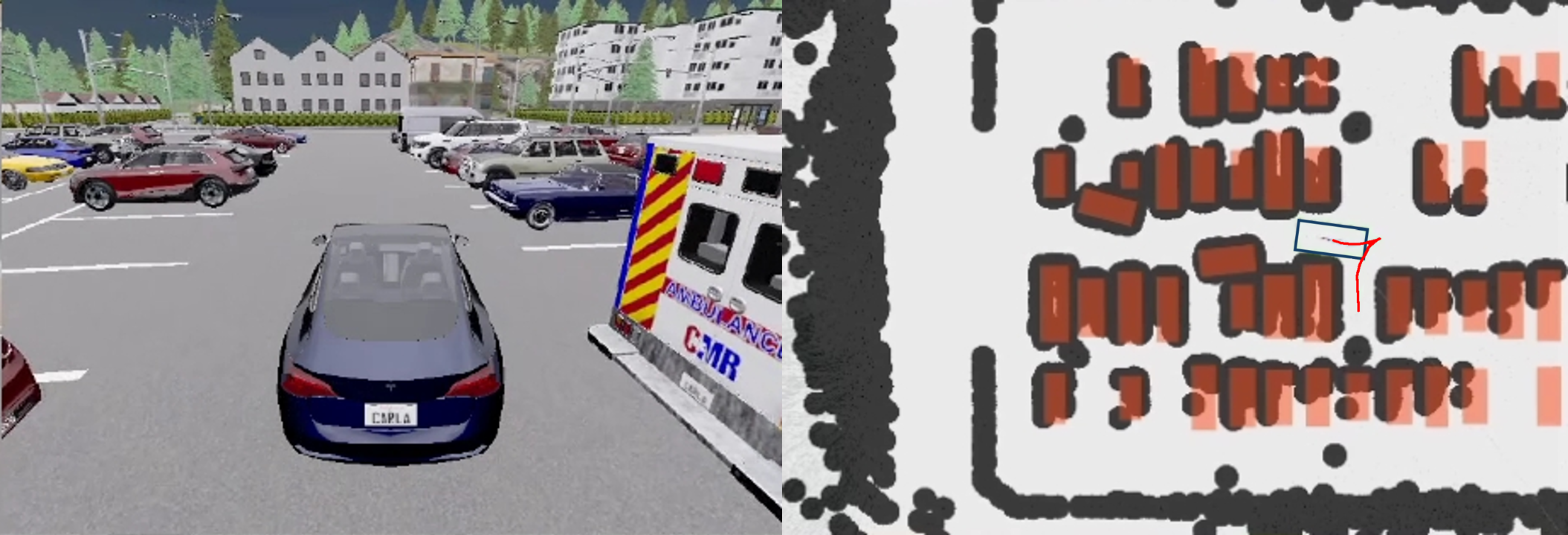}}
  \end{minipage}

  \caption{Autonomous parking in a cluttered parking lot. The maneuver comprises two phases: forward driving toward the target parking space, followed by reverse parking into the designated slot. (a) shows the executed vehicle trajectory. (b), (c) show representative third-person views (left) and corresponding top-down views (right); red rectangles denote ground-truth vehicle bounding boxes, and the boundary edges extracted from these boxes are used as PSDF inputs.}
  \label{fig:carla_traj}
\end{figure*}

Two further experiments assess onboard execution and the applicability of PSDF-MPC beyond differential-drive platforms.

PSDF-MPC was deployed on a Scout Mini and executed on its onboard Jetson Xavier in an unstructured environment. VLP-16 LiDAR measurements were processed using the same projection, line-fitting, and safety-inflation pipeline as in the Gazebo experiment. On the Jetson Xavier, the mean and P95 per-step computation times were 21.9 and 42.5 ms, respectively. These results indicate that most control cycles were completed within the 100-ms sampling period. Fig.~\ref{fig:real_traj}(a) shows the driving environment and executed trajectory, while Fig.~\ref{fig:real_traj}(b) and (c) show the perceived obstacle boundaries.

The parking experiment was conducted in CARLA \cite{dosovitskiy2017carla} by replacing the differential-drive prediction model with an Ackermann kinematic model. A Hybrid-A* planner \cite{dolgov2008practical} generated a kinematically feasible reference trajectory involving both forward and reverse motion. Ground-truth bounding boxes of surrounding vehicles were converted into polygonal edge sets and supplied to PSDF. The controller tracked the reference while enforcing the same PSDF-based safety constraints. Fig.~\ref{fig:carla_traj} shows collision-free forward driving and reverse parking in a dense parking lot.

These results support the applicability of PSDF-MPC to onboard navigation and Ackermann parking while retaining the same PSDF-based safety constraint and its local linearization.

\section{CONCLUSION}
\label{conclusion}

This paper presented PSDF, a training-free formulation that computes exact signed distances between a convex polygonal robot footprint and convex obstacle components and returns their minimum. Its branch-free, piecewise-differentiable tensor structure supports batched GPU evaluation, with state gradients obtained through automatic differentiation. PSDF-MPC incorporates these values and gradients into SQP-RTI through one locally linearized safety constraint per prediction stage. The resulting sparse QP is solved on the CPU, with its size and sparsity determined by system dimensions and horizon length rather than obstacle and edge counts. Query benchmarks showed favorable scaling with obstacle count and batch size over the tested range, highlighting the efficiency of value and gradient evaluation for larger batched workloads. Closed-loop experiments showed high navigation success rates and mean and P95 computation times below the sampling period, supporting the real-time feasibility of PSDF-MPC for collision-free navigation in dense environments. Onboard deployment and Ackermann parking simulations further demonstrated the applicability of PSDF-MPC to navigation on real robotic systems and to vehicles with different kinematics.

The formulation is restricted to planar polygonal geometry, and the current implementation and experiments hold the edge set fixed over each prediction horizon. Future work includes extending the proposed method to three-dimensional models and time-varying obstacles, and incorporating safety margins that account for perception uncertainty, while preserving the same batched interface.

\bibliographystyle{Bibliography/IEEEtranTIE}
\bibliography{bib/refs}

\begingroup
\makeatletter
\def\@IEEEBIOskipN{1\baselineskip}
\makeatother

\begin{IEEEbiography}[{\includegraphics[width=1in,height=1.25in,clip,keepaspectratio]{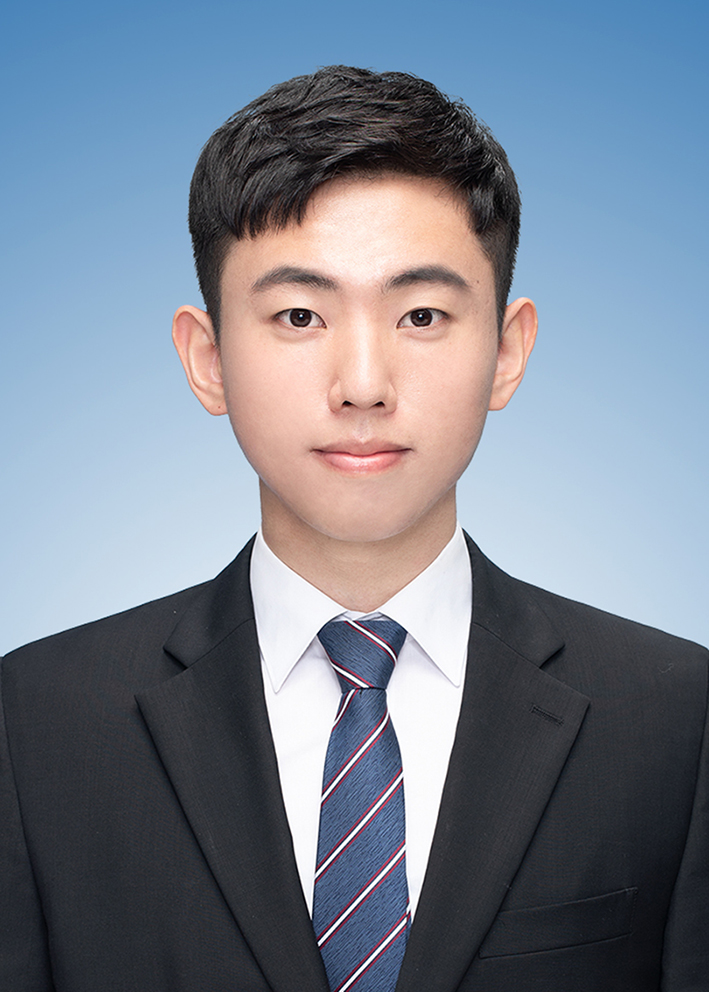}}]{Taekwon Ga}
received the B.S. degree in Electronic and Electrical Engineering from Sungkyunkwan University, Suwon, South Korea, in 2020. He is currently a Ph.D. candidate in the Department of Mechanical Engineering at Yonsei University, working in the field of machine learning and control systems. His research interests include optimal control of mobile robots, machine learning, and robotics.
\end{IEEEbiography}

\begin{IEEEbiography}
[{\includegraphics[width=1in,height=1.25in,clip,keepaspectratio]{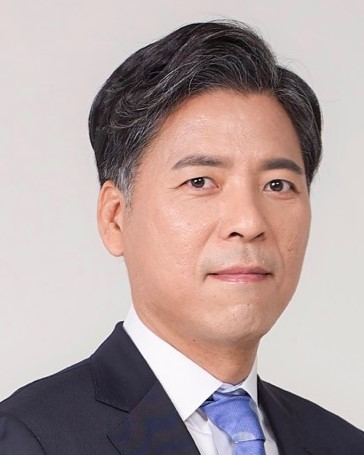}}]{Jongeun Choi}
received his B.S. degree in Mechanical Design and Production Engineering from Yonsei University, Seoul, Republic of Korea, in 1998, and his M.S. and Ph.D. degrees in Mechanical Engineering from the University of California, Berkeley, in 2002 and 2006, respectively. He is currently a Professor at the School of Mechanical Engineering, Yonsei University, where he has also been affiliated with the Department of Artificial Intelligence since 2020. In 2023, he served as a visiting scholar at UC Berkeley.

Before joining Yonsei University, Dr. Choi was a faculty member at Michigan State University, where he worked as an Assistant Professor (2006--2012) and later as an Associate Professor (2012--2016) in the Departments of Mechanical Engineering and Electrical and Computer Engineering.

Dr. Choi's research interests include machine learning, systems and control, reinforcement learning, and Bayesian methods, with applications to robotics, autonomous driving, human-robot interaction, and AI in healthcare. He has been serving as a Senior Editor for the International Journal of Control, Automation, and Systems since 2023 and previously served as an Associate Editor for the IEEE Robotics and Automation Letters in 2018, the Journal of Dynamic Systems, Measurement, and Control (2014--2019), and the International Journal of Precision Engineering and Manufacturing (2017--2018).

Dr. Choi has received several awards, including the Best Paper Award at the RSS 2023 Workshop on Symmetries in Robot Learning and the 12th International Conference on Ubiquitous Robots and Ambient Intelligence in 2015. His papers were finalists for the Best Student Paper Awards at the 24th American Control Conference in 2005 and the Dynamic Systems and Control Conference in 2011 and 2012. He was also a recipient of the NSF CAREER Award in 2009. Dr. Choi is an ASME Fellow and a member of IEEE.
\end{IEEEbiography}
\endgroup

\end{document}